\documentclass[10pt,letterpaper,twocolumn]{article}
\pdfoutput=1

\usepackage[letterpaper,top=48pt,bottom=48pt,left=49pt,right=49pt,
            columnsep=18pt,footskip=28pt]{geometry}
\usepackage{times}
\usepackage{amsmath,amssymb}
\usepackage{graphicx}
\usepackage{booktabs}
\usepackage[font=small,labelfont=bf,skip=5pt]{caption}
\usepackage{enumitem}
\usepackage[T1]{fontenc}
\usepackage[utf8]{inputenc}
\usepackage{cite}
\usepackage{xcolor}
\definecolor{linkblue}{rgb}{0.08235,0.28235,0.69804}
\usepackage[colorlinks=true,linkcolor=linkblue,citecolor=linkblue,
            urlcolor=linkblue,pdftitle={Toward Uncertainty-Aware, Physics-Grounded Human World Models},
            pdfauthor={Yufei Zhang, Zijun Cui}]{hyperref}

\usepackage{titlesec}
\usepackage{titling}
\titleformat{\section}{\normalfont\Large\bfseries\raggedright}{\thesection}{1em}{}
\titleformat{\subsection}{\normalfont\large\bfseries\raggedright}{\thesubsection}{1em}{}
\titlespacing*{\section}{0pt}{13pt plus 2pt minus 2pt}{6pt}
\titlespacing*{\subsection}{0pt}{11pt plus 2pt minus 1pt}{7pt}
\title{{\fontsize{14.35}{17}\selectfont\bfseries Toward Uncertainty-Aware, Physics-Grounded Human World Models}}

\author{Yufei Zhang\\
Independent Researcher\\
{\tt\small yufeizhang96@outlook.com}
\and
Zijun Cui\\
Michigan State University\\
{\tt\small cuizijun@msu.edu}
}

\date{}

\begin{document}
\maketitle

\section*{Abstract}

A robot that acts around people needs a world model of the human: a belief over
the person's body state and a dynamics model that predicts where that state goes
next. This abstract argues that two published components form exactly these two
halves, and specifies how to couple them. KNOWN~\cite{known} reconstructs the 3D
human body while estimating both aleatoric and epistemic uncertainty in real time
(11.6\,ms per frame); its aleatoric term identifies unreliable observations, its
epistemic term identifies inputs from low-density regions of the training data,
and refinement driven by the epistemic term reduces the error on such inputs from
81.3 to 70.3\,mm. PhysMoP~\cite{physmop} predicts future body configuration by
solving the Euler--Lagrange equations of the body with learned dynamics terms,
reducing the 80\,ms prediction error from 9.6 to 2.1\,mm on Human3.6M and from
12.1 to 0.7\,mm on 3DPW. We define the interface between the two, propose three
mechanisms by which KNOWN's uncertainty enters PhysMoP, and specify an evaluation
protocol on in-the-wild video. No new experimental results are reported here.

\section{Introduction}

A learned world model earns its place in a robot by predicting the consequences of
actions before they are taken, and by knowing where its predictions cannot be
trusted: a planner is an optimizer and will exploit model error wherever the model
is silent about it~\cite{mopo,morel}. In human environments the most consequential
part of the world state is a person, an articulated system with 72 rotational
degrees of freedom in the SMPL parameterization~\cite{smpl}, observed through depth
ambiguity and frequent occlusion, and with a collision costing far more than a
delay.

Neither the belief nor the dynamics model exists today in a form a robot can use. Human mesh estimators return a
point estimate, or a distribution over plausible bodies that does not separate
observation noise from model ignorance~\cite{prohmr,biggs,glopro}; certifiable
human-robot collaboration pipelines add a separate out-of-distribution detector to
the pose estimator because predicted aleatoric uncertainty does not capture
unfamiliar inputs~\cite{thumm2026}. Human motion predictors are trained and
evaluated on motion-capture histories~\cite{physmop,simlpe,eqmotion}; how they
behave on estimated input has not been tested. Physical structure is what keeps a
prediction plausible once its input is estimated rather than measured~\cite{physpt},
but so far it has been used to refine the past, not to carry an uncertain state
forward.

Two published components supply the two halves. KNOWN~\cite{known} reconstructs the
3D body and quantifies both aleatoric and epistemic uncertainty of the
reconstruction from a single forward pass, in 11.6\,ms. PhysMoP~\cite{physmop}
predicts future body configuration by embedding the Euler--Lagrange equations into
the predictor, with sub-centimeter root-aligned error at 80--160\,ms, the horizons
on which a robot's control loop operates. Both are defined on SMPL pose parameters,
and this abstract contributes:

\begin{enumerate}[leftmargin=*,itemsep=0.6pt,topsep=1pt]
  \item KNOWN as the belief over the body state and PhysMoP as its dynamics
        model, together with the interface that would join the two
        (Sec.~\ref{sec:interface});
  \item three potential mechanisms for that belief to enter the predictor:
        initial-condition sampling, epistemic gating, and smoothing weighted by
        the aleatoric covariance (Sec.~\ref{sec:mechanisms});
  \item a possible protocol for evaluating them (Sec.~\ref{sec:protocol}).
\end{enumerate}

\begin{figure*}[t]
  \centering
  {\small (a) KNOWN: belief over body state with aleatoric and epistemic uncertainty}\\[2pt]
  \includegraphics[width=434pt]{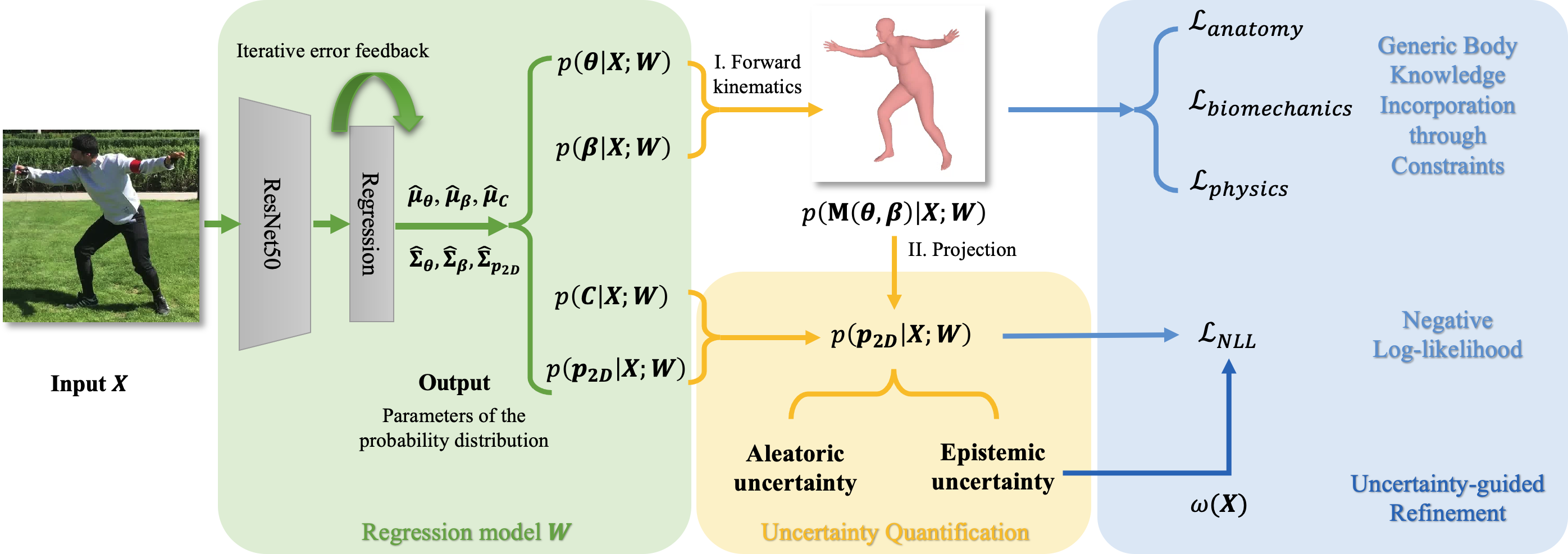}\\[4pt]
  {\small $\downarrow$\, per-frame belief $p(Y\mid X_t)\;\rightarrow\;$ history $q_{1:T}$
   (Sec.~\ref{sec:coupling})}\\[4pt]
  \includegraphics[width=434pt,trim=19.2 19.2 40.8 25.0,clip]{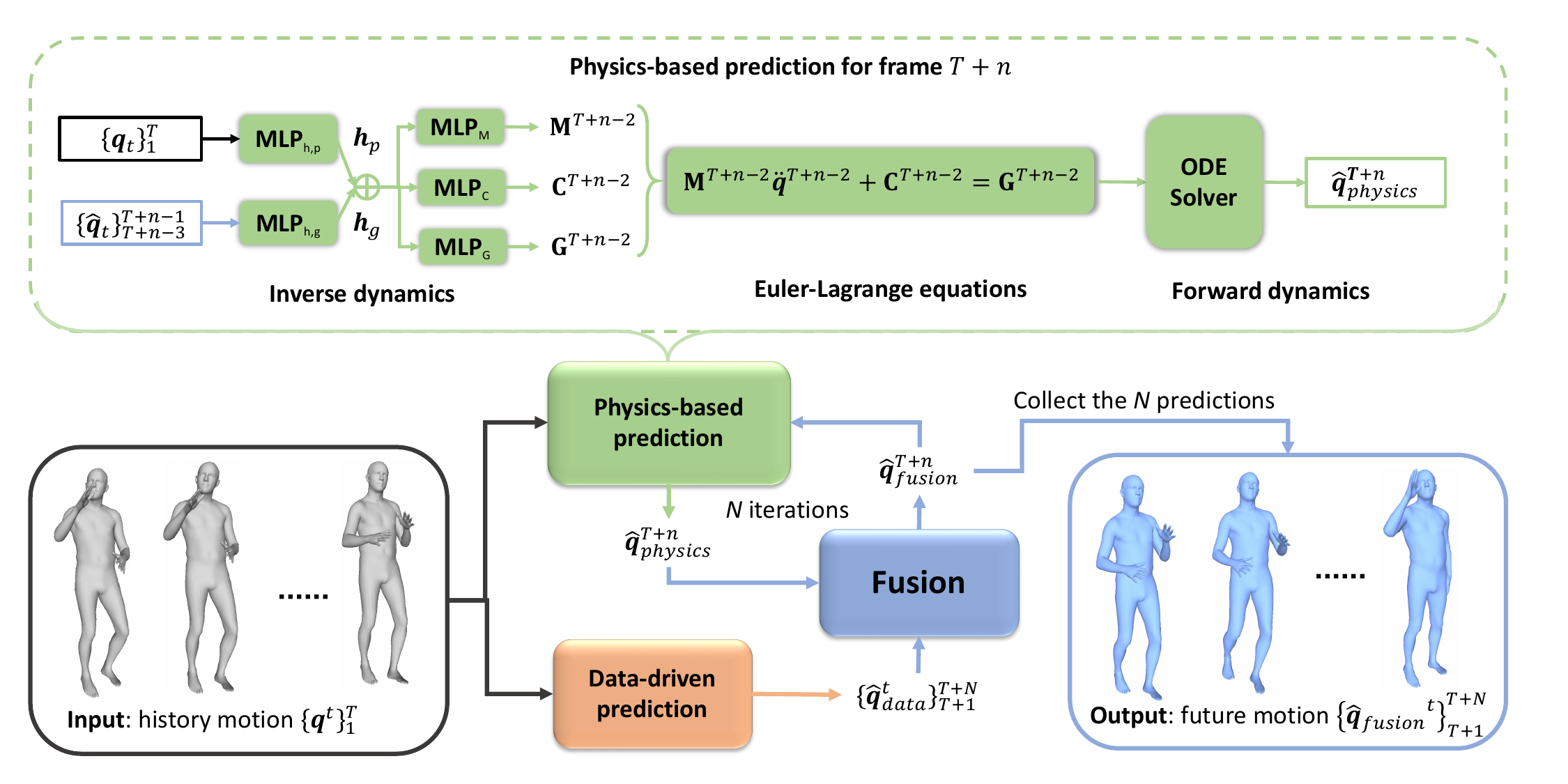}\\[2pt]
  {\small (b) PhysMoP: physics-grounded prediction of future body configuration}
  \caption{The two halves of the human world model. (a) KNOWN~\cite{known}: a
  regression network outputs a distribution over SMPL pose $\theta$, shape $\beta$
  and camera $C$; forward kinematics and projection give the observation model
  $p(\mathbf{p}_{2D}\mid X)$, from which aleatoric and epistemic uncertainty are
  read off; the uncertainties enter training through the NLL and
  uncertainty-guided refinement. (b) PhysMoP~\cite{physmop}: a physics branch
  predicts the Euler--Lagrange terms $M$, $C$, $G$ and integrates them, a data
  branch predicts the whole future, and a fusion network combines the two. The
  arrow is the coupling proposed in Sec.~\ref{sec:coupling}: KNOWN's per-frame
  belief becomes the history $q_{1:T}$ that PhysMoP integrates forward.}
  \label{fig:overview}
\end{figure*}

\section{KNOWN: Uncertainty-Aware \mbox{Reconstruction}}
\label{sec:known}

\subsection{Uncertainty Modeling}

KNOWN is a single two-stage probabilistic network (Fig.~\ref{fig:overview}a).
Given an image $X$, the first stage predicts a distribution over the body state
$Y=(\theta,\beta)$, the SMPL pose and shape,
\begin{equation}
  p(Y \mid X) = \mathcal{N}\!\bigl(\mu_Y(X),\, \Sigma_Y(X)\bigr),
\end{equation}
with a diagonal $\Sigma_Y$, together with weak-perspective camera parameters
$C(X)$. The second stage is an observation model: for a sample $Y$, forward
kinematics $J(\cdot)$ and projection $\Pi(\cdot)$ give the mean 2D keypoints, and
the network predicts an observation covariance,
\begin{equation}
  p(\mathbf{p} \mid Y, X) = \mathcal{N}\!\bigl(\Pi(J(Y),C),\, \Sigma_{\mathbf{p}}(X)\bigr).
\end{equation}
The law of total variance splits the covariance of the observation into two terms,
\begin{equation}
  \mathrm{Cov}[\mathbf{p}\mid X] =
  \underbrace{\Sigma_{\mathbf{p}}(X)}_{\text{aleatoric}} +
  \underbrace{\mathrm{Cov}_{Y\sim p(Y\mid X)}\!\bigl[\Pi(J(Y),C)\bigr]}_{\text{epistemic}},
  \label{eq:totvar}
\end{equation}
mirroring the structure of~\cite{kendall}, with the predicted state distribution in
the role of the posterior. Applied to the mesh, $\mathrm{Cov}_Y[V(Y)]$ over vertices
$V$ gives a per-vertex 3D uncertainty, and the joint regressor turns it into a
$3\times3$ position covariance at every joint.

Two properties matter here. First, the uncertainty is defined on a
structured, physically constrained state rather than on pixels or a latent, and it
propagates through kinematics: a rotation perturbation at a joint displaces every
descendant, so position covariance accumulates along the kinematic chain
(Fig.~\ref{fig:known}: larger circles at hands and feet). Second, both terms come
from one forward pass of the backbone, since the covariance in Eq.~\eqref{eq:totvar}
is a Monte-Carlo estimate over samples of $Y$ that pass only through the SMPL
forward function and the projection. KNOWN runs in 11.6\,ms with 103.9\,MB of
parameters (Table~\ref{tab:known}), against 16.4\,ms and 223.0\,MB for the
normalizing-flow model ProHMR~\cite{prohmr}, and 566.6\,MB for the multi-hypothesis
model of Biggs et al.~\cite{biggs}.

\subsection{Uncertainty-Guided Training}

Both uncertainties act during training. The aleatoric term enters through the data
term, a negative log-likelihood under the observation model,
\begin{equation}
  \begin{aligned}
    \mathcal{L}_{\mathrm{NLL}} &= \mathbb{E}_{Y}\!\Bigl[
      \mathbf{r}(Y)^{\top}\Sigma_{\mathbf{p}}^{-1} \mathbf{r}(Y)
      + \log\det\Sigma_{\mathbf{p}}\Bigr],\\
    \mathbf{r}(Y) &= \hat{\mathbf{p}} - \Pi(J(Y),C),
  \end{aligned}
\end{equation}
so a keypoint with a large predicted covariance contributes little to the residual
term instead of being fit; annotations pooled from datasets with differing keypoint
definitions and quality are absorbed rather than fit. The epistemic term enters
through a refinement stage in which the loss is re-weighted toward inputs whose
predicted state distribution is widest, so the model is trained hardest where its
own belief is least certain.

\subsection{Results}

\begin{figure}[t]
  \centering
  \includegraphics[width=198.4pt]{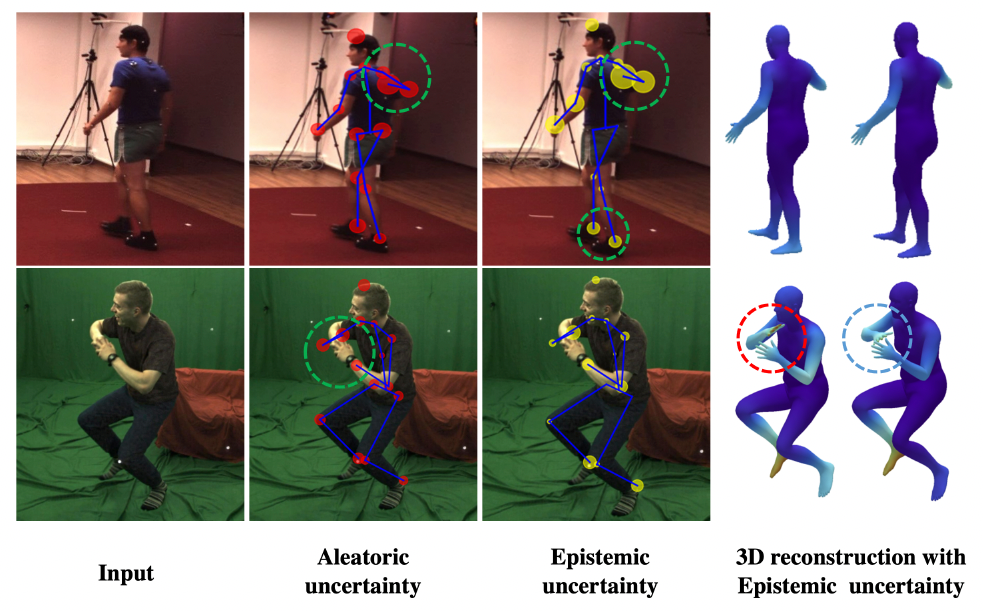}
  \caption{Qualitative results of KNOWN on two Human3.6M images~\cite{known}. Left
  to right: input; per-joint aleatoric uncertainty (radius of the red circles);
  per-joint epistemic uncertainty (yellow); reconstruction colored by per-vertex
  epistemic uncertainty, before (left) and after (right) uncertainty-guided
  refinement. Dashed circles mark the occluded or ambiguous parts the uncertainties
  single out.}
  \label{fig:known}
\end{figure}

In~\cite{known}, the training images with the largest aleatoric uncertainty are
those with self-occlusion, poor image quality and annotation errors, including
left--right swaps; Fig.~\ref{fig:known} shows the per-joint version, with the
largest circles on the occluded arm. The epistemic term is inversely related to
training-data density: the images with the highest epistemic uncertainty, the
``minorities'' defined by a percentile threshold on the training set, are those with
large camera angles, severe occlusion and extreme poses, the conditions a
robot-mounted camera produces. Refinement driven by the epistemic term reduces the
error on minority images from 81.3 to 70.3\,mm; on that subset KNOWN outperforms
HMR~\cite{hmr} (107.3\,mm) and matches SPIN~\cite{spin} (71.0\,mm).

\begin{table}[t]
  \centering
  \footnotesize
  \setlength{\tabcolsep}{5pt}
  \begin{tabular}{@{}lcccc@{}}
    \toprule
    Method & All & Minority & MB & ms \\
    \midrule
    HMR~\cite{hmr}            & --   & 107.3 & --    & --   \\
    SPIN~\cite{spin}          & --   & 71.0  & --    & --   \\
    ProHMR~\cite{prohmr}      & --   & --    & 223.0 & 16.4 \\
    Biggs et al.~\cite{biggs} & -- & -- & 566.6 & --  \\
    \midrule
    KNOWN, before refinement  & 58.1 & 81.3  & 103.9 & 11.6 \\
    KNOWN                     & \textbf{55.9} & \textbf{70.3} & \textbf{103.9} & \textbf{11.6} \\
    \bottomrule
  \end{tabular}
  \caption{KNOWN on Human3.6M: mean per-joint position error after Procrustes
  alignment (P-MPE, mm) on all test images and on the minority subset, with model
  size and per-frame latency, from~\cite{known}. ``Minority'': test images whose
  epistemic uncertainty exceeds a percentile threshold set on the training set. HMR
  and SPIN are evaluated on the same subset.}
  \label{tab:known}
\end{table}

\begin{figure}[t]
  \centering
  \includegraphics[width=183.5pt]{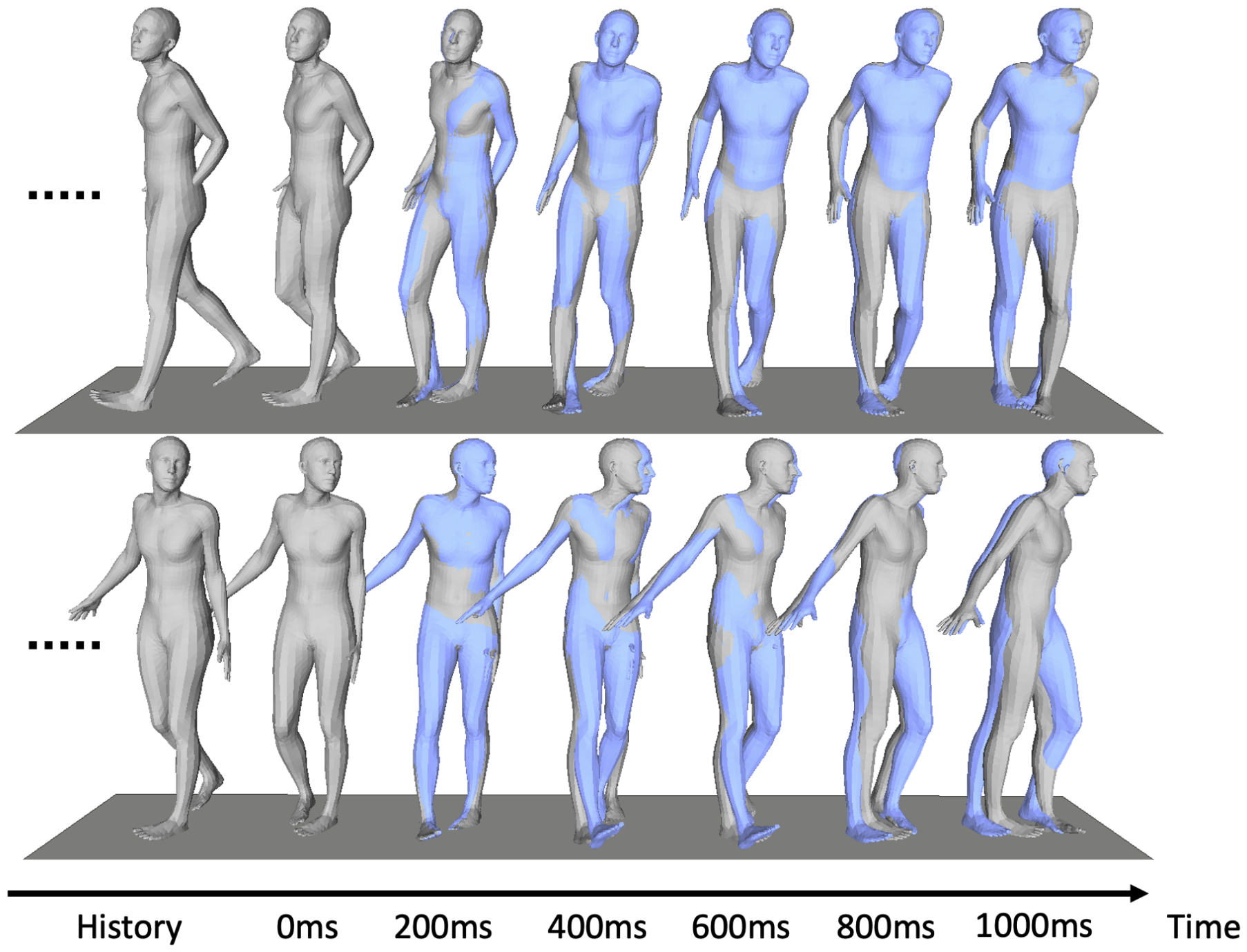}
  \caption{Qualitative results of PhysMoP on two Human3.6M test
  sequences~\cite{physmop}: ground truth in grey, prediction in blue, from the last
  history frame to 1\,s ahead. Larger overlap means smaller prediction error.}
  \label{fig:physmop}
\end{figure}

\begin{table}[t]
  \centering
  \footnotesize
  \setlength{\tabcolsep}{4pt}
  \begin{tabular}{@{}llccccc@{}}
    \toprule
    Data & Method & 80\,ms & 160 & 320 & 400 & 1000 \\
    \midrule
    H3.6M & siMLPe~\cite{simlpe}        & 9.6 & 21.7 & 46.3 & 57.3 & 109.4 \\
    H3.6M & EqMotion~\cite{eqmotion}    & 9.1 & 20.1 & 43.7 & 55.0 & 106.9 \\
    H3.6M & PhysMoP, physics branch     & \textbf{2.1} & \textbf{7.6} & \textbf{28.4} & 42.2 & 148.2 \\
    H3.6M & PhysMoP, data branch        & 4.4 & 10.6 & 33.8 & 47.8 & \textbf{103.3} \\
    H3.6M & PhysMoP                     & \textbf{2.1} & \textbf{7.6} & \textbf{28.4} & 43.8 & 103.9 \\
    \midrule
    3DPW  & siMLPe~\cite{simlpe}        & 12.1 & 22.1 & 38.1 & 44.5 & 72.2 \\
    3DPW  & PhysMoP                     & \textbf{0.7} & \textbf{3.6} & \textbf{15.1} & \textbf{19.8} & \textbf{70.9} \\
    \bottomrule
  \end{tabular}
  \caption{PhysMoP motion prediction: root-aligned MPJPE (mm) at each horizon,
  from~\cite{physmop}. Human3.6M follows protocol P1; the 3DPW rows use a model
  trained on AMASS and evaluated cross-dataset. Inputs are ground-truth
  motion-capture histories in all rows.}
  \label{tab:physmop}
\end{table}

\section{PhysMoP: Physics-Grounded Motion Prediction}
\label{sec:physmop}

\subsection{Model}

PhysMoP treats the body as a physical system. Its generalized coordinate is
$q=(\theta,T)\in\mathbb{R}^{75}$, the joint angles of the 24 SMPL joints as Euler
angles and the root translation, with the rotation order of each joint chosen from
its biomechanical range. The body obeys the Euler--Lagrange equations
\begin{equation}
  M_t\,\ddot{q}_t + C_t = G_t,
\end{equation}
where $M_t$ is the generalized inertia, $C_t$ collects Coriolis, centrifugal and
gravitational terms, and $G_t$ the generalized forces, internal and external. The
model predicts these quantities: features of the whole history
$h_p=\mathrm{MLP}(q_{1:T})$ and of the three most recent estimates
$h_g=\mathrm{MLP}(\hat{q}_{t-3:t-1})$ feed three heads that output $M$, $C$ and $G$;
forward dynamics, $\ddot{q}_t = M_t^{-1}(G_t-C_t)$ followed by Euler integration,
yields the next configuration. Applied iteratively this is a physics-based rollout
that accumulates error, so a data branch in the style of siMLPe~\cite{simlpe}
predicts the whole future from the history, and a fusion network outputs per-frame
weights $\hat{w}_t$ combining the two, regularized toward the physics branch.

\subsection{Results}

Table~\ref{tab:physmop} reports root-aligned MPJPE. On Human3.6M the 80\,ms error
drops from 9.6\,mm (siMLPe) and 9.1\,mm (EqMotion~\cite{eqmotion}) to 2.1\,mm; on
3DPW, with a model trained on AMASS~\cite{amass} and never fine-tuned, from 12.1 to
0.7\,mm. The ablation rows explain the design: the physics branch alone is the most
precise model up to 320\,ms and the worst at 1\,s (148.2\,mm against 103.3\,mm for
the data branch), the compounding error of an autoregressive rollout, and the
learned fusion keeps the short-horizon precision of physics with the long-horizon
accuracy of data. On Human3.6M the acceleration error falls from
8.2\,mm/frame$^2$ for the data branch to 2.3 for the fused model.
Fig.~\ref{fig:physmop} shows predicted meshes against ground truth over 1\,s: the
overlap is near-complete up to 400\,ms, and the deviation at 800--1000\,ms is the
compounding error of Table~\ref{tab:physmop}.

\section{Coupling KNOWN and PhysMoP}
\label{sec:coupling}

Every number in Table~\ref{tab:physmop} was obtained from motion-capture ground
truth, whereas a robot has only KNOWN's per-frame belief to work from.

\subsection{Interface}
\label{sec:interface}

Per frame, KNOWN provides a mean and covariance over $\theta$ and $\beta$ and,
through sampling, a $3\times3$ covariance per joint. PhysMoP consumes a
$T{=}25$-frame history of $q$. Rotations are converted from KNOWN's rotation
parameterization to PhysMoP's Euler angles under the shared rotation-order
convention, and histories are resampled to PhysMoP's training frame rate. Root
translation is not observable under KNOWN's weak-perspective camera; as in recent
certifiable human-robot collaboration systems that take root depth from a depth
sensor~\cite{thumm2026}, it is supplied externally. Shape is averaged over the
sequence.

\subsection{Three Mechanisms}
\label{sec:mechanisms}

The three follow KNOWN's own decomposition of Eq.~\eqref{eq:totvar}: the first
uses the belief as a whole, the second the aleatoric term, the third the
epistemic term.

\emph{The whole belief: initial-condition sampling.} Sample $S$ histories from the
per-frame beliefs, run PhysMoP on each, and read off the mean and spread of the
predicted futures. This needs no retraining and turns the deterministic predictor
into a predictive distribution whose width is inherited from perception.
In an analogous pipeline, propagating pose-estimation covariance into a motion
predictor gave prediction sets 11$\times$ smaller than ISO 13855 constant-velocity
sets at comparable coverage~\cite{thumm2026}. Sampling frames independently ignores
temporal correlation of estimation error and is conservative.

\emph{The aleatoric term: covariance-weighted smoothing.} The physics branch conditions on
three consecutive frames, the minimum needed to specify velocity and acceleration.
Any acceleration estimate formed from three noisy frames inherits that noise
through a second difference, which amplifies it sharply at short frame intervals,
so per-frame estimation error is the quantity this branch is most sensitive to. A
linear Gaussian smoother in $q$-space, with a constant-acceleration process model
and KNOWN's $\Sigma_Y$ as time-varying measurement covariance, discounts frames with
wide beliefs before they are differenced; the covariance carries the needed
information, since it is largest on exactly the occluded and ambiguous parts where
the estimate is least reliable (Fig.~\ref{fig:known}). The aleatoric term enters the
same smoother as a per-frame inflation of $\Sigma_Y$ proportional to the trace of
$\Sigma_{\mathbf{p}}$, a heuristic choice: poorly observed keypoints yield a less
reliable state regardless of the parameter covariance.

\emph{The epistemic term: gating.} Frames whose epistemic uncertainty exceeds the
minority threshold are, on KNOWN's own evidence, inputs from low-density regions of
its training distribution: selected by epistemic uncertainty alone, they carry 26\%
higher error than the test-set average even after refinement (70.3 vs.\ 55.9\,mm,
Table~\ref{tab:known}), so recent-state derivatives should not be trusted there. The
simplest policy widens the measurement covariance in proportion to the epistemic
term; a stricter one withholds the physics branch and falls back to the data branch,
which is what PhysMoP's fusion weight already expresses. Where recent safety
pipelines attach a separate out-of-distribution detector to the
estimator~\cite{thumm2026}, here the estimator produces the signal itself, per joint
and per frame.

\subsection{Evaluation Protocol}
\label{sec:protocol}

3DPW~\cite{3dpw} test sequences are the benchmark: in-the-wild video with SMPL
ground truth, used for cross-dataset evaluation in both original papers.
Ground-truth camera extrinsics map KNOWN's estimate into the world frame. Five
inputs are compared at 80, 160, 320 and 400\,ms by root-aligned MPJPE: (1)
ground-truth histories, the upper bound of Table~\ref{tab:physmop}; (2)
point-estimate histories; (3) sampled histories; (4) covariance-weighted smoothing;
(5) smoothing with epistemic gating. The physics and data branches are reported
separately, and results are split by the minority criterion. Either outcome is informative: if the
physics branch degrades faster than the data branch under estimated input,
structured world models need calibrated state uncertainty, and (3)--(5) measure how
much of the gap it recovers; if not, physical structure itself regularizes state
noise. A calibration check completes the protocol: the spread of (3) is compared
with the error of (2), and the empirical coverage of its 90\% interval is reported
per horizon.

\section{Discussion}

\emph{Which representation.} A structured state with a known decoder
(SMPL kinematics and projection) allows the total-variance decomposition to be
computed by sampling through the decoder only, and allows the Euler--Lagrange
structure to be imposed on the dynamics; neither is available in a pixel-space or
an unstructured-latent world model.

\emph{Two uncertainties, two decisions.} In model-based RL, epistemic
uncertainty is penalized~\cite{mopo} or sought~\cite{plan2explore}; aleatoric
uncertainty is what a policy must be robust to. The same split separates a filter
from a monitor here, and KNOWN's refinement stage is that loop closed offline.

\emph{Predictor as world model.} Up to action conditioning, the human
component of a robot's world model is a motion predictor supplying the future
occupancy that speed-and-separation monitoring~\cite{thumm2026} and manipulation
safety shields~\cite{thumm2022} consume. Offline model-based RL penalizes
ensemble-based epistemic uncertainty of the dynamics~\cite{li2025}; human mesh
trackers propagate observation uncertainty through a learned motion
model~\cite{glopro}; neither treats the perception belief as the initial condition
of a physics-structured dynamics model.

\emph{Limitations.} KNOWN's epistemic term is induced by the predicted
state distribution and validated empirically, not derived from a posterior over
network weights. KNOWN provides no metric root translation, and the protocol assumes
a known camera pose. Neither component is conditioned on the robot's actions, so the
composite forecasts rather than answers counterfactuals; PhysMoP's fusion weights
are not yet informed by the perception uncertainty itself.

{\footnotesize
\bibliographystyle{ieeetr}
\bibliography{refs}
}

\end{document}